\documentclass[11pt]{article}

\usepackage{fullpage}
\usepackage[ruled, linesnumbered, vlined, commentsnumbered]{algorithm2e}
\usepackage{graphicx,subfig} % figure related
\usepackage{amsfonts,amssymb,amsmath,amsthm,amsopn,mathtools}	% math related
\usepackage{booktabs,diagbox,colortbl,multirow,tabularx,threeparttable,hhline}
\usepackage[listings,skins,breakable]{tcolorbox}

\usepackage{enumerate}
\usepackage{authblk}
\usepackage{footnote}
\usepackage{hyperref}
\usepackage{prettyref}
\usepackage{cite}
\usepackage{setspace}
\usepackage{color}
\usepackage{xcolor}  % Required for custom colors
\usepackage{scrpage2}
\usepackage{geometry}

\usepackage{tikz}
\usepackage{pgfplots}
\usetikzlibrary{positioning,shapes,shadows,arrows,calc}
\tikzstyle{component}=[rectangle, draw=black, rounded corners, fill=blue!40, drop shadow, text centered, anchor=north, text=white, minimum height=1cm]
\tikzstyle{arrow}=[->, thick]

\pgfplotsset{compat=1.12}
\usetikzlibrary{intersections}
\usetikzlibrary{pgfplots.statistics}
\usepgfplotslibrary{fillbetween}

\definecolor{myblue}{RGB}{34,31,217}
\definecolor{mycyan}{gray}{.7}
\definecolor{Gray}{gray}{0.9}

\newtheorem{remark}{Remark}

\DeclareMathOperator*{\argmax}{argmax}
\DeclareMathOperator*{\argmin}{argmin}

\def\our{\texttt{BiLO-Auto-TSF/ML}}

\newcommand{\pref}{\prettyref}

\newrefformat{fig}{Fig.~\ref{#1}}
\newrefformat{tab}{Table~\ref{#1}}
\newrefformat{sec}{Section~\ref{#1}}
\newrefformat{alg}{Algorithm~\ref{#1}}
\newrefformat{property}{Property~\ref{#1}}
\newrefformat{theorem}{Theorem~\ref{#1}}
\newrefformat{definition}{Definition~\ref{#1}}
\newrefformat{corollary}{Corollary~\ref{#1}}
\newrefformat{lemma}{Lemma~\ref{#1}}
\newrefformat{conj}{Conjecture~\ref{#1}}
\newrefformat{def}{Definition~\ref{#1}}
\newrefformat{eq}{equation~(\ref{#1})}
\newrefformat{app}{Appendix~\ref{#1}}

\usepackage{lscape}

\begin{document}

%% title
\title{\vspace{-1ex}\LARGE\textbf{Automated Few-Shot Time Series Forecasting based on Bi-level Programming}\footnote{This manuscript is submitted for potential publication. Reviewers can use this version in peer review.}}

%% authors and affiliations
\author[]{\normalsize Jiangjiao Xu}
\author[]{\normalsize Ke Li}
\affil[]{\normalsize Department of Computer Science, University of Exeter, EX4 4QF, Exeter, UK}
%\affil[1,2]{\normalsize ABC, City, Nationality}
%\affil[1]{\normalsize Email: \texttt{j.xu@exeter.ac.uk}}
\affil[]{\normalsize Email: \texttt{k.li@exeter.ac.uk}}

\date{}
\maketitle

\vspace{-3ex}
{\normalsize\textbf{Abstract: } 
New micro-grid design with renewable energy sources and battery storage systems can help improve greenhouse gas emissions and reduce the operational cost. To provide an effective short-/long-term forecasting of both energy generation and load demand, time series predictive modeling has been one of the key tools to guide the optimal decision-making for planning and operation. One of the critical challenges of time series renewable energy forecasting is the lack of historical data to train an adequate predictive model. Moreover, the performance of a machine learning model is sensitive to the choice of its corresponding hyperparameters. Bearing these considerations in mind, this paper develops a BiLO-Auto-TSF/ML framework that automates the optimal design of a few-shot learning pipeline from a bi-level programming perspective. Specifically, the lower-level meta-learning helps boost the base-learner to mitigate the small data challenge while the hyperparameter optimization at the upper level proactively searches for the optimal hyperparameter configurations for both base- and meta-learners. Note that the proposed framework is so general that any off-the-shelf machine learning method can be used in a plug-in manner. Comprehensive experiments fully demonstrate the effectiveness of our proposed BiLO-Auto-TSF/ML framework to search for a high-performance few-shot learning pipeline for various energy sources.
}


% The following tex files can be different according to your unique case.
%!TeX root=main.tex

\section{Introduction}
\label{sec:introduction}

Smart grid technology has become the driving force that enables an effective management and distribution of renewable energy sources such as solar, wind and hydrogen. It connects a variety of distributed energy resource assets to the power grid. The relationship between the smart grid and renewable energy revolves around gathering data. With flourishing developments of the Internet of things (IoT), utility companies are able to quickly detect and resolve service issues through continuous self-assessments and self-healing by leveraging heterogeneous and time series data collected on the smart grid. In particular, time series predictive modeling, which provides short- and/or long-term forecasting of both energy generation and load demand, has been one of the key tools to guide optimal decision-making for planning and operation of utility companies without over/underestimating the capabilities of renewable energy infrastructures~\cite{AmjadyKZ10}.

As discussed in some recent survey papers (e.g.,~\cite{AkhterMMM19,NatarajanK19,LaiCCP20}), there have been a noticeable amount of efforts on applying machine learning methods for time series renewable energy forecasting in the past two decades and beyond. For example, neural networks (NNs) with multi-layer perceptron were developed to forecast the daily solar radiation in time series dataset by using a transfer function of hidden layers~\cite{PaoliVMN10}. Recurrent neural networks (RNNs) such as long short-term memory (LSTM)~\cite{KongDJHXZ19}, which take historical time series data as the input and predict the trajectory over a certain time horizon, have been proved to be effective because they consider both the instantaneous interactions within contiguous time steps and the long-term dependencies stored in memory cells. In~\cite{MajidpourQCGP15}, a $K$-nearest neighbor based time weighted dot product dissimilarity measure was proposed to improve the forecasting accuracy and reduce the processing time. In~\cite{ChenCL04}, support vector regression (SVR) was applied to make a mid-term time series load prediction. In~\cite{KouGG13}, a sparse online warped Gaussian process regression was developed to provide a short-term probabilistic prediction of wind power generation with a non-Gaussian predictive distribution.

To promote the further uptake of machine learning in smart grid industry, we need to address two imperative challenges.
\begin{itemize}
    \item\underline{\textit{Small data challenge}}: Fitting a time series model with an adequate statistical confidence usually requires sufficient data. Unfortunately, this is hardly met in real-life scenarios. For example, when planning an island grid, due to various technical problems such as equipment failures, measurement errors or restrictions on the daily power supply time~\cite{McLarty17}, there may exist substantial gaps in the collected historical power consumption data, i.e., incomplete or missing data. Even worse, some islands may not have any historical data at all due to the laggard in infrastructure. All these significantly compromise the prediction performance in time series forecasting. How to use the limited historical data of multiple islands to predict the energy of an island without historical data is a big challenge for conventional prediction models.
%The scenario considered in this article is when one of the islands does not have enough historical data while the other ones have only partial historical data. 
    \item\underline{\textit{Hyperparameter tuning challenge}}: The performance of most, if not all, machine learning methods are very sensitive to their hyperparameters such as neural architectures in NNs, kernel functions and regularization methods in SVR. The optimal choice of hyperparameters also vary across tasks and datasets~\cite{ZophL17}. The black-box nature of machine learning models leads to a considerable barrier to domain experts, who are interested in applying machine learning methods yet have sufficient time and/or resources to learn the inside out, for choosing the optimal hyperparameters to achieve the state-of-the-art performance.
\end{itemize}

Bearing these challenges in mind, this paper develops \our, a first of its kind bi-level programming framework to automate the time series forecasting in smart grid with limited historical data. The key contributions are summarized as follow.
\begin{itemize}
    \item To address the small data challenge, we propose to use meta-learning to improve the generation performance of the base-learner to unseen tasks by only referring to limited historical data. Although there have been some successful applications of meta-learning in various domains such as object detection~\cite{SnellSZ17,Perez-RuaZHX20}, landmark prediction~\cite{GuiWRM18}, and image/video generation~\cite{GordenBBNT19,WangLTLKC19}, it has rarely been explored in the context of smart grid, to the best of our knowledge.
      
    \item Due to the nested structure between base- and meta-learners, there exist complex interactions among their associated hyperparameters. In this paper, we hypothesize that it may not be able to achieve a peak performance if the hyperparameters associated with the base- and meta-learners are set independently. Although there have been some previous attempts on hyperparameter optimization for meta-learning, they mainly consider the ones associated with the meta-learner, such as the learning rate in the inner loop~\cite{Vanschoren18}. In this paper, our ambition is to automate the optimal design of a few-shot learning\footnote{In this paper, we use meta-learning and few-shot learning interchangeably.} pipeline from a bi-level programming perspective.
    
    \item In \our, the upper-level optimization searches for the optimal hyperparameter settings associated with both base- and meta-learners by a Monte Carlo tree search (MCTS)~\cite{BrownePWLCRTPSC12}; while the lower-level optimization implements a model-agnostic gradient-based meta-learning as done in~\cite{Finnal17}. In other words, any off-the-shelf machine learning can be used in \our\ in a plug-in manner.
    
    \item To validate the effectiveness of our proposed \our\ framework, we consider some selected real-world energy forecasting tasks for smart grid infrastructure planning in island areas at the English Channel. In particular, three prevalent machine learning methods are used as the base-learners for a proof-of-concept purpose. Extensive experimental results fully demonstrate the effectiveness of our proposed \our\ framework for time series forecasting with highly limited historical data.
\end{itemize}

The rest of this paper is organized as follows. \pref{sec:related} provides a pragmatic overview of some related works. \pref{sec:proposed} delineates the implementation of our proposed \our\ framework. \pref{sec:setup} gives the experimental setup while the empirical results are presented and analyzed in~\pref{sec:results}. Finally, \pref{sec:conclusion} concludes this paper and threads some lights on potential future directions.

\section{Related Works}
\label{sec:related}

This section provides a pragmatic overview of some selected developments of both time series forecasting in smart grid and hyperparameter optimization for meta-learning.

\subsection{Time Series Forecasting in Smart Grid}
\label{sec:related_sg}

Electric load demand or renewable energy generation forecasting is critical to the operation and management of a smart grid. Based on the length of time period, the predictive modeling can be divided into short-, mid-, and long-term forecasting~\cite{ZhengXZL17}. In recent years, NNs have been widely applied to obtain latent information to build prediction models, e.g., dynamic choice artificial neural network model~\cite{WangLSZ16}, generalized regression neural network~\cite{LiW12}, and nonlinear autoregressive neural network models with exogenous input (NARX)~\cite{BuitragoA17}. Nevertheless, NNs are notorious for overfitting and are also suffered from local optima in the backpropagation~\cite{ArifJANGF20}. To mitigate the overfitting problem by raising a new data dimension, Shi et al.~\cite{ShiXL18} proposed to use a pooling-based deep recurrent neural network for short-term household load forecasting. Moon et al.~\cite{MoonJRRH20} proposed to synthesize more than one deep neural network model with multiple hidden layers to select a model with the best prediction performance. In~\cite{Aly20}, Aly proposed a hybrid of wavelet neural network and Kalman filter for short-term load forecasting problems. Mid-term forecasting is used to coordinate load dispatch and balance load demand and renewable energy generation~\cite{KhuntiaRM16}. Jiang et al.~\cite{WeiHLHH19} proposed a dynamic Bayes network for mid-term forecasting problem to predict the peak time load in a year. In~\cite{GrzegorzPS21}, Grzegorz et al. proposed a hybrid deep learning model for mid-term forecasting that combined exponential smoothing (ETS), multi-layer LSTM and ensemble learning which have shown competitive performance against other models like ARIMA and ETS. Long-term forecasting is used to predict the power consumption and generation ranging from a few years to a couple of decades for the system planning and expansion in a smart grid. In~\cite{AgrawalMT18} and~\cite{ZhengXZL17}, variants of RNN have been proposed for long-term time series load forecasting and have shown better results than other forecasting methods like NARX and SVR.

\subsection{Hyperparameter Optimization for Meta-Learning}
\label{sec:related_m}

Meta-learning has been proven to be effective in multiple tasks scenarios where task-agnostic knowledge is extracted from the same distribution of tasks with small datasets and used to find good starting parameters of the baseline machine learning model for new tasks~\cite{HospedalesAMS20}. It has been widely appreciated that the performance of machine learning is sensitive to the choice of the corresponding hyperparameters. For example, The existing gradient-based meta-learning usually rely on the choice of an appropriate optimizer to fine tune the parameters of the meta-learner. Furthermore, there are various hyperparameters associated with the base-learner such as the neural architecture and the learning rate, the configuration of which can influence the predictive performance. In the past decade, there have been many efforts devoated to the hyperparameter optimization for meta-learning. For example, in order to reduce the sensitivity to the hyperparameters, Li et al.~\cite{LiZCL17} proposed to use a stochastic gradient descent method to update the inner-loop learning rate for the meta-learner. The experimental results have shown the effectiveness of hyperparameter optimization for the meta-learner. In~\cite{AntoniouES18}, Antoniou et al. proposed an improved gradient-based meta-learning method in which the inner-loop learning rate is updated according to the performance of the selected model. In~\cite{RusuRSVPOH18}, Rusu et al. proposed a latent embedding optimization approach to optimize a range of hyperparameters in meta-learning. The experimental results have shown superior performance against some gradient-based hyperparameter adaptation. Franceschi et al.~\cite{FranceschiFSGP18} proposed a bi-level programming framework to optimize the parameters of a neural network along with the hyperparameters of a meta learner in a concurrent manner. In~\cite{BaikCCKL20}, Baik et al. proposed a fast adaptation approach to predict the adaptive hyperparameters by using the current parameters and their gradients. The proposed method from a random initialization with an adaptive learning of hyperparameters outperform other existing algorithms.

\begin{remark}
    Most, if not all, existing time series forecasting models in the smart grid literature require sufficient amount of historical data. Unfortunately, this can be hardly met in real-world applications of which available data is usually scarce especially for the power system design and planning in remote areas like an isolated island network.
\end{remark}

\begin{remark}
    From the above literature review, we find that the existing studies on hyperparameter optimization only take the hyperparameters in the inner loop of meta-learning into consideration. However, it is not difficult to envisage that there exist certain dependencies between the hyperparameters associated with both base- and meta-learners. Unfortunately, the concurrent optimization w.r.t. these two types of hyperparameters have been largely ignored in the literature.
    % As discussed above, many hyperparameter optimization algorithms only consider to reduce the inner-loop hyperparameter sensitivity to improve the performance. The important insight of machine learning hyperparameter optimization is that there are dependencies between different hyperparameters, which affects the search for the optimal model in the meta-learning literature.
\end{remark}

%!TeX root=main.tex

\section{Automated Few-Shot Learning for Time Series Forecasting}
\label{sec:proposed}

This section delineates the implementation of our proposed \our\ framework that automates the design of a few-shot learning pipeline for time series forecasting with limited data from a bi-level programming perspective. We start with the overarching bi-level programming problem formulation considered in this paper. Then, we delineate the algorithmic implementation of the optimization routines at both levels respectively.

\subsection{Problem Formulation of Bi-level Programming}
\label{sec:problem_formulation}

\begin{figure*}[t!]
    \centering
    \includegraphics [width=\linewidth]{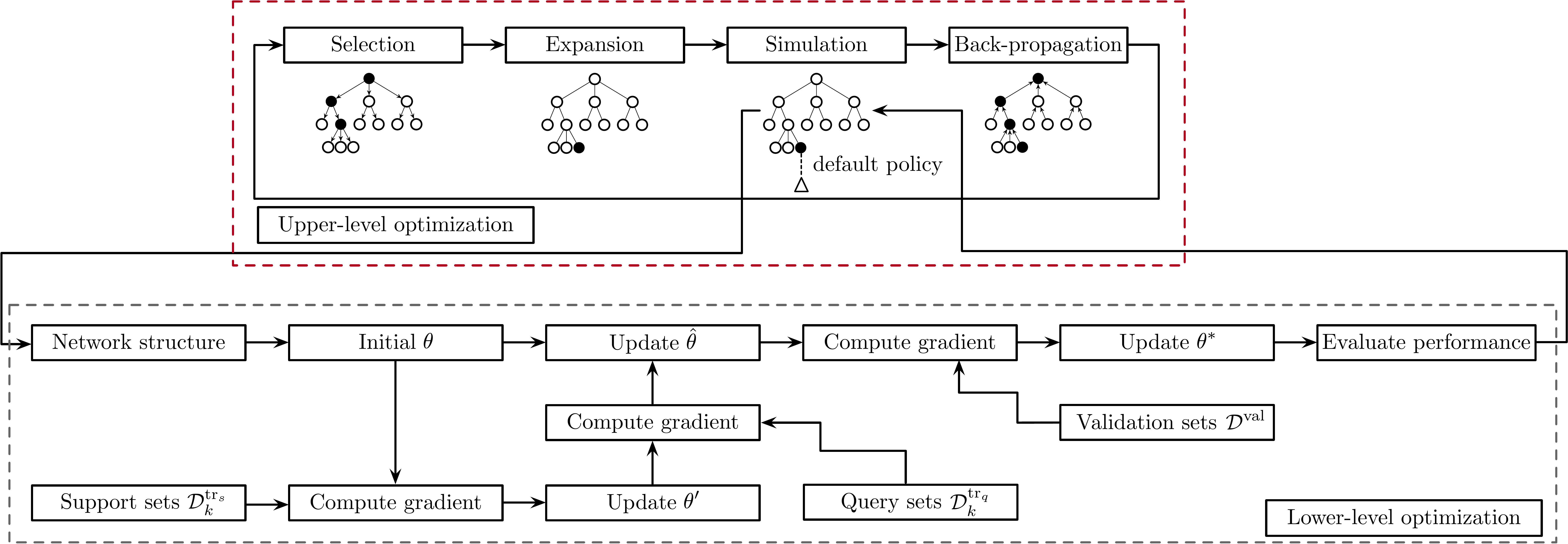}
    \caption{The overall architecture and workflow of \our.}
    \label{fig:architecture}
\end{figure*}

The \our\ framework involves a sequence of $l\geq 1$ decisions that choose the hyperparameters associated with the baseline machine learning model along with its parameters, the learning rates of both inner and outer loop of the meta-learning, the optimization algorithms for empirical loss function optimization and the number of shots in meta-learning, respectively. At the $i$-th decision step ($1\leq i\leq l$), a design component $c_i\in\mathcal{C}_i$ is selected where $\mathcal{C}_i$ is a finite set of possible alternative options at the $i$-th decision step. The space of hyperparameters associated with $c_i$ is denoted as $\Theta(c_i)$ and a complete pipeline structure is an $\ell$-tuple $\mathbf{c}=(c_1,\cdots,c_l)\in\mathcal{C}=\mathcal{C}_1\times\cdots\times\mathcal{C}_l$ where $\Theta(\mathbf{c})=\Theta(c_1)\times\cdots\times\Theta(c_l)$ is its associated hyperparameter space.

The key challenge of the automated design of a few-shot learning pipeline for time series forecasting considered in this paper is an intertwined optimization of the parametric optimization of the associated hyperparameters $\mathbf{c}$ and the gradient-based optimization associated with the meta-learning simultaneously. In this paper, we propose to formulate this intertwined optimization problem as the following bi-level programming problem:
\begin{equation}
    \begin{aligned}
        &\mathrm{minimize}\quad\mathcal{L}^\mathtt{val}(\mathbf{c},\hat{\boldsymbol{\theta}};\mathcal{D}^\mathtt{val})\\
        &\mathrm{subject\ to} \quad \hat{\boldsymbol{\theta}}:=\argmin_{\boldsymbol{\theta}}\mathcal{L}_\mathtt{meta}^\mathtt{tr}(\mathbf{c},\boldsymbol{\theta};\mathcal{D}^\mathtt{tr})
    \end{aligned},
    \label{eq:blp}
\end{equation}
where $\mathcal{L}^\mathtt{val}$ and $\mathcal{L}^\mathtt{tr}_\mathtt{meta}$ denote the upper- and lower-level objective functions, and $\mathbf{c}\in\Theta(\mathbf{c})$ and $\hat{\boldsymbol{\theta}}\in\mathbb{R}^m$ denote the upper- and lower-level variables. More specifically, at the upper level, we aim to identify the best few-shot learning pipeline $\mathbf{c}^\ast$ having the optimal hyperparameters along with the optimal parameters $\hat{\boldsymbol{\theta}}^\ast$ associated with the baseline machine learning model that minimizes $\mathcal{L}^\mathtt{val}$ (i.e., the validation empirical risk) at the end of meta-training where:
\begin{equation}
    \mathcal{L}^\mathtt{val}(\mathbf{c},\hat{\boldsymbol{\theta}};\mathcal{D}^\mathtt{val})\triangleq\frac{1}{|\mathcal{D}^\mathtt{val}|}\sum_{(\mathbf{x}^i,y^i)\in\mathcal{D}^\mathtt{val}}\ell_\mathbf{c}^\mathtt{val}\Big(y^i,f(\mathbf{x}^i;\hat{\boldsymbol{\theta}})\Big),
\end{equation}
where $\ell_\mathbf{c}^\mathtt{val}\Big(y^i,f(\mathbf{x}^i;\hat{\boldsymbol{\theta}})\Big)$ is the loss function regarding a given few-shot learning pipeline $\mathbf{c}$ on the validation set. Note that this validation empirical risk requires the parameters of the baseline machine learning model to be optimized by a meta-learning process at the lower level. The training empirical risk associated with the lower-level meta-learning is defined as:
\begin{equation}
    \mathcal{L}^\mathtt{tr}_\mathtt{meta}\triangleq\sum_{\mathcal{D}^\mathtt{tr}_k\sim\mathcal{D}^\mathtt{tr}}\mathcal{L}^\mathtt{tr}(\mathbf{c},\boldsymbol{\theta};\mathcal{D}^\mathtt{tr}_k),
\end{equation}
where
\begin{equation}
    \mathcal{L}^\mathtt{tr}(\mathbf{c},\boldsymbol{\theta};\mathcal{D}^\mathtt{tr}_k)\triangleq\frac{1}{|\mathcal{D}^\mathtt{tr}_k|}\sum_{(\mathbf{x}^i,y^i)\in\mathcal{D}^\mathtt{tr}_k}\ell_\mathbf{c}^\mathtt{tr}\Big(y^i,f(\mathbf{x}^i;\boldsymbol{\theta})\Big),
\end{equation}
where $\ell_\mathbf{c}^\mathtt{tr}\Big(y^i,f(\mathbf{x}^i;\boldsymbol{\theta})\Big)$ is the loss function regarding $\mathbf{c}$ on the meta-training set. 

The overall architecture of \our\ is given in~\pref{fig:architecture}. It consists of two nested levels of optimization routines. Given the discrete nature of the upper-level hyperparameter optimization problem, \our\ applies a Monte-Carlo tree search (MCTS), as delineated in~\pref{sec:upper}, to explore the hyperparameter space thus gradually navigating the exploration towards some most promising regions within the search tree. As for the lower-level optimization, we develop a gradient-based meta-learning approach as in~\cite{Finnal17}. As detailed in~\pref{sec:lower}, it identifies the promising initial parameters for the baseline machine learning model to adapt to the target task(s) with limited data.

\subsection{Upper-level Optimization Routine}
\label{sec:upper}

%Suitable hyperparameter combinations can effectively improve the initialization parameters of the machine learning model and reduce the prediction error~\cite{LiXCT20}. The upper-level optimization considers the use of a tree-structured to search the hyperparameter space and iteratively explores the most promising areas.
In view of the discrete nature of the search space of hyperparameters, this paper considers using a tree-based search to implement the hyperparameter optimization at the upper level. As shown in~\pref{fig:architecture}, the MCTS algorithm iterates over the following four steps.
\begin{itemize}
    \item\underline{\textit{Selection}}: The MCTS algorithm starts from the root node $n_\mathtt{r}$ and recursively selects internal nodes downward until it reaches a leaf node $n_\mathtt{l}$. In particular, the tree-path from $n_\mathtt{r}$ to an internal node represents a partial solution, i.e., an incomplete pipeline. As for each parent (non-leaf) node $n_\mathtt{p}$, its child node $n_\mathtt{c}\in\mathcal{N}_\mathtt{c}$ is selected by maximizing the upper confidence bound for trees~\cite{Auer03}:
\begin{equation}
    n_c:=\argmax_{n_c\in\mathcal{N}_\mathtt{c}} \Bigg\{\frac{Q(n_\mathtt{c})}{\mathbb{N}(n_\mathtt{c})}+\alpha\sqrt{\frac{\ln \mathbb{N}(n_\mathtt{p})}{\mathbb{N}(n_\mathtt{c})}}\Bigg\},
\end{equation}
where $Q(n_\mathtt{c})$ is the expected reward for $n_\mathtt{c}$, $\mathbb{N}(n_\mathtt{c})$ and $\mathbb{N}(n_\mathtt{p})$ is the number of times that $n_\mathtt{c}$ and $n_\mathtt{p}$ has been visited, respectively, and $\alpha$ is a parameter that controls the trade-off between exploration and exploitation.

	\item\underline{\textit{Expansion}}: Once a leaf node is reached, one or more child nodes will be amended to expand the tree structure. In this paper, we apply the rapid action value estimation method~\cite{GellyS11} to select the new nodes. The number of child nodes associated with its parent node $n_\mathtt{p}$ is selected and constructed according to a progressive widening trick~\cite{AugerCT13}. In particular, a new child node can be amended if and only if the $\lceil \mathbb{N}(n_\mathtt{p})^\kappa\rceil$ increases by one, where $0<\kappa<1$ is a constant coefficient.

    \item\underline{\textit{Simulation}}: If the expansion step finishes whereas a terminal node is not yet reached (i.e., we come up with an incomplete few-shot learning pipeline), a default policy (we use a random sampling strategy in this paper) is applied to select the remaining options until it reaches a terminal node. Thereafter, given the generated few-shot learning pipeline $\mathbf{c}$, its reward $Q(\mathbf{c})$ is evaluated as the validation empirical risk. In particular, the parameters of the baseline machine learning model are optimized by a meta-learning process at the lower level.

%        Once the expansion step is completed, an incompleted pipeline structure $\mathbf{c_k}=(c_1,\cdots,c_k), k \leq \ell$ will be determined. Afterward, a default policy (i.e., a random sampling strategy) will be implemented to generate a full pipeline structure, including the selected node $n_{t+1}$ to the terminal node $n_T$. Given the completed meta-learning pipeline structure, the reward value $Q_{v}$ corresponding to the validation empirical risk will be evaluated by a meta-learning process at the lower-level.

    \item\underline{\textit{Back-propagation}}: The reward value obtained in the expansion step is back-propagated to update the $\mathbb{N}$ and $Q$ values associated with all nodes along the visited tree-path until the root node. In particular, for each visited node $n$, the update rule is as follows:
        \begin{equation}%{4}
            \begin{aligned}
                \mathbb{N}(n) &:= \mathbb{N}(n)+1, \\
                Q(n) &:= Q(n)+Q(\mathbf{c}).
            \end{aligned}
        \end{equation}

\end{itemize}

Note that the above four steps iterate until either the computational budget is exhausted.
%\textit{\textbf{Stopping criterion:}} When the computational budget is exhausted or a completed pipeline structure is determined, then the upper-level MCTS will stop.

%The pseudo-code of MCTS at the upper-level optimization is shown in~\pref{alg:MCTS}. It can produce an asymptotic optimality guarantees with sufficient exploitation. Note that the decision made at the beginning of the root node is the choice of the learning algorithm and the default policy at the simulation step is the choice of the random sampling strategy in this paper.
%\begin{algorithm}[t]
%    \caption{Upper-level optimization to find the best hyperparameter settings}
%    \label{alg:MCTS}
%    \KwIn{Overall hyperparameter space}
%    \KwOut{Optimized machine learning model}
%
%    \textbf{Create} a root node $n_0$;\\
%    \For{$i \leftarrow 1$ \KwTo $N$}{
%        Descend the tree to a leaf node $n_t$ ;\\
%        Choose a child (action) from $n_t$ ;\\
%        Run a simulation from $n_{t+1}$ to the terminal node ;\\
%        Obatin a reward from the lower-level optimization;\\
%        Backpropagation to update the visited nodes;\\
%    }
%    \Return Best meta-learning pipeline $\mathbf{c}^\ast$
%\end{algorithm}

\subsection{Lower-level Optimization Routine}
\label{sec:lower}

As discussed in Sections~\ref{sec:problem_formulation} and~\ref{sec:upper}, the lower-level optimization aims to train a machine learning model with limited historical data. In the following paragraphs, we will elaborate on the general setup of our meta-learning along with the meta-training process.
% from other islands to predict the data for a new island (no historical data or only few limited data). In this subsection, we will define the meta-learning setup and present the general form of lower-level optimization in a few-shot learning scenario.
\begin{itemize}
    \item\underline{\textit{Meta-learning setup}}: As introduced in~\pref{sec:introduction}, when considering the planning of island grids, a key challenge is the shortage of historical data that render the classic machine learning pipeline largely ineffective. This motivates us to use meta-learning to mitigate this small data challenge. Formally, let us denote the data collected from $T$ islands as $\mathcal{T}:=\{\mathcal{T}_i\}_{i=1}^T$ to constitute the training dataset. In particular, we assume that $\mathcal{T}$ is sampled from a fixed distribution $\mathbb{P}(\mathcal{T})$. For each island, the corresponding training dataset is constituted as $\mathcal{D}^\mathtt{tr}_i:=\mathcal{D}^{\mathtt{tr}_s}_i \bigcup\mathcal{D}^{\mathtt{tr}_q}_i$ where $i\in\{1,\cdots,T\}$, $\mathcal{D}^{{\mathtt{tr}_s}}_i$ and $\mathcal{D}^{{\mathtt{tr}_q}}_i$ denote the support and query sets respectively. In this paper, we have $|\mathcal{D}^{\mathtt{tr}_s}_i|:=|\mathcal{D}^{\mathtt{tr}}_i|\times 80\%$.

  	\item\underline{\textit{Meta-learning process}}: Inspired by~\cite{Finnal17}, we develop a gradient-based meta-learning process where the parameters $\boldsymbol{\theta}$ of the baseline machine learning are updated as:
  	    \begin{equation}
        \boldsymbol{\theta}^{\prime}:=\boldsymbol{\theta}-\alpha \nabla_{\boldsymbol{\theta}}\mathcal{L}^\mathtt{tr}(\mathbf{c},\boldsymbol{\theta};\mathcal{D}^\mathtt{tr_s}_k),
        \label{eq:fogt}
        \end{equation}
        where $\alpha$ is the inner-loop learning rate and $\nabla_{\boldsymbol{\theta}}\mathcal{L}^\mathtt{tr}(\mathbf{c},\boldsymbol{\theta};\mathcal{D}^\mathtt{tr_s}_k)$ is the gradient of the loss function regarding the support set $\mathcal{D}^\mathtt{tr_s}_k$ where $k$ is for all picked up islands. 
        \begin{itemize}
            \item During the meta-training phase, we randomly pick up $N\leq T$ islands, each of which contains $K$ data instances drawn from the corresponding island to constitute $\mathcal{D}_k^\mathtt{tr}$ where $k\in\{1,\cdots,N\}$. the model parameters are updated by minimizing the loss function associated with $\boldsymbol{\theta}$ across the query sets of $N$ islands sampled from $\mathbb{P}(\mathcal{T})$ defined as:
            \begin{equation}
            \mathcal{L}^\mathtt{tr}_\mathtt{meta}(\mathbf{c},\boldsymbol{\theta}^{\prime};\mathcal{D}^\mathtt{tr_q}_k)\triangleq\sum_{\mathcal{D}^\mathtt{tr_q}_k\sim\mathcal{D}^\mathtt{tr_q}}\mathcal{L}^\mathtt{tr}(\mathbf{c},\boldsymbol{\theta}^{\prime};\mathcal{D}^\mathtt{tr_q}_k),
            \label{eq:fogopt}
            \end{equation}
            where \pref{eq:fogopt} is calculated by using the updated parameters $\boldsymbol{\theta}^{\prime}$ for the given query sets $\mathcal{D}^\mathtt{tr_q}$ and the loss function regarding the $k$-th island is defined as:
            \begin{equation}
            \mathcal{L}^\mathtt{tr}(\mathbf{c},\boldsymbol{\theta}^\prime;\mathcal{D}^\mathtt{tr_q}_k):=\sum_{(\mathbf{x}^{i},y^{i})\in \mathcal{D}^\mathtt{tr_q}_k}||f(\mathbf{x}^{i};\boldsymbol{\theta}^\prime)-y^{i}||^2_2.
            \label{eq:lossfunc}
            \end{equation}
            Accordingly, the parameters $\hat{\boldsymbol{\theta}}$ of the baseline machine learning model are updated as:
            \begin{equation}
            \hat{\boldsymbol{\theta}}:=\boldsymbol{\theta}-\beta \nabla_{\boldsymbol{\theta}^{}}\mathcal{L}^\mathtt{tr}_\mathtt{meta}(\mathbf{c},\boldsymbol{\theta}^{\prime};\mathcal{D}^\mathtt{tr_q}_k),
            \label{eq:foga}
            \end{equation}
            where $\beta$ is the outer-loop learning rate and $\hat{\boldsymbol{\theta}}$ is the updated parameters of the baseline machine learning model after a one-step gradient descent.%This process is repeated until the end of the iteration.
            
            \item After the meta-training phase, the updated parameters $\hat{\boldsymbol{\theta}}$ will be fine-tuned on the validation sets $\mathcal{D}^\mathtt{val}$ with a few gradient descent steps as:
            %we obtain the final model parameters $\hat{\boldsymbol{\theta}}$. When the representation is suitable to new islands, we only need to simply modify the parameters slightly to reduce the prediction error. Given the validation data sets $\mathcal{D}^\mathtt{val}$, we can fine-tune the model with limited steps of gradient descent. The gradients to optimize the model parameters can be formulated as:
            \begin{equation}
            \boldsymbol{\theta}^\ast:=\hat{\boldsymbol{\theta}}-\gamma \nabla_{\hat{\boldsymbol{\theta}}} \mathcal{L}^\mathtt{val}_\mathtt{meta}(\mathbf{c},\hat{\boldsymbol{\theta}};\mathcal{D}^\mathtt{val}),
            \label{eq:ftu}
            \end{equation}
            where $\gamma$ is the validation learning rate.
        \end{itemize}
        
\end{itemize}

At the end, $\boldsymbol{\theta}^\ast$ will be used as the optimal parameters of the baseline machine learning model for predicting the unseen data collected from a new island. The workflow of the lower-level optimization is illustrated at lower-level optimization part of~\pref{fig:architecture} and the pseudo-code is given in~\pref{alg:BML}.

%   	We are inspired by a meta-learning technique to learn the parameters of the machine learning models associated with limited data which are more transferable than the models learned by other techniques~\cite{Finnal17}. A gradient-based method is used to update the model parameters that can adapt to improve the prediction performance for a new island. When an island $\mathcal{T}_k$ with respect to dataset $\mathcal{D}^\mathtt{tr}_k$ is sampled to train the machine learning model, the model parameters $\boldsymbol{\theta}$ will update to $\boldsymbol{\theta}^{\prime}$. The update rule based on the gradient descent can be given as follows:

\begin{algorithm}[t!]
    \caption{Model-agnostic gradient-based meta-learning at the lower level of \our.}
    \label{alg:BML}
    \KwIn{Meta training data set $\mathcal{D}^\mathtt{tr}=(\mathcal{D}^{\mathtt{tr}_s},\mathcal{D}^{\mathtt{tr}_q})$, meta validation data set $\mathcal{D}^\mathtt{val}$, parameters of base learner and meta learner including $\alpha$, $\beta$, $\gamma$}
    \KwOut{Predicted outputs}
    \textbf{Initialize} model parameter $\boldsymbol{\theta}$;\\
    \For{$iteration \leftarrow 1,2,...$ \KwTo}{
        Sample $N$ islands randomly from $T$;\\
        \For{all  $k={1,\cdots,N}$ \KwTo}{
        Sample support sets $\mathcal{D}^{\mathtt{tr}_s}_k$ ; \\
        Compute the loss $\mathcal{L}^\mathtt{tr}(\mathbf{c},\boldsymbol{\theta};\mathcal{D}^\mathtt{tr_s}_k)$; \\
        Obtain the $\boldsymbol{\theta}^{\prime}$ via gradient descent in~(\ref{eq:fogt}) ;\\
        Sample query sets $\mathcal{D}^{\mathtt{tr}_q}_k$;\\
        Compute the loss $\mathcal{L}^\mathtt{tr}_\mathtt{meta}(\mathbf{c},\boldsymbol{\theta}^{\prime};\mathcal{D}^\mathtt{tr_q}_k)$; \\
        }
    Obtain $\hat{\boldsymbol{\theta}}$ via gradient descent using $\mathcal{L}^\mathtt{tr}_\mathtt{meta}(\mathbf{c},\boldsymbol{\theta}^{\prime};\mathcal{D}^\mathtt{tr_q}_k)$ with respect to $\boldsymbol{\theta}$; \\
    }
    
    Obtain the parameters $\boldsymbol{\theta}^*$ using $\mathcal{D}^\mathtt{val}$ via~(\ref{eq:ftu}) with a couple of gradient descent steps;\\
    
    Feed data into the machine learning model to predict the outputs for new island;

    \Return Prediction results
\end{algorithm}

%!TeX root=main.tex

\section{Experimental Setup}
\label{sec:setup}

This section introduces the setup of our empirical study including the dataset, parameter settings, the performance metric, and the statistical tests~\cite{ChenLY18,ZouJYZZL19,LiZZL09,BillingsleyLMMG19,LiZLZL09,Li19,LiK14,LiFK11,LiKWTM13,CaoKWL12,CaoKWL14,LiDZZ17,LiKD15,LiKWCR12,LiWKC13,CaoKWLLK15,LiDY18,WuKZLWL15,LiKCLZS12,LiDAY17,LiDZ15,LiXT19,GaoNL19,LiuLC19,LiZ19,KumarBCLB18,CaoWKL11,LiX0WT20,LiuLC20,LiXCT20,WangYLK21,ShanL21,LaiL021,LiLLM21,WuKJLZ17,LiCSY19,LiLDMY20,WuLKZ20,PruvostDLL020}.
\begin{itemize}
    \item\underline{\textit{Dataset}}: Our empirical study considers the energy forecasting tasks for smart grid infrastructure planning in islands at the English Channel~\cite{MatthewFCWTHMAYH18}. We consider three different energy sources including the wind generation, the photovoltaic (PV) generation, and the load demand. The training set consists of data collected from four islands including Ushant, Molene, Sein and Isles of Scilly while those obtained from the island Lundy constitute the validation set. As discussed in~\pref{sec:introduction}, the key challenge here is the lack of sufficient historical data. In particular, for each energy source, there is only a week's time series data for each island.

    \item\underline{\textit{Parameter settings}}: There are $\ell=5$ hyperparameters considered in our automated few-shot learning pipeline. 
        \begin{itemize}
            \item Note that our proposed \our\ is model agnostic where any off-the-shelf machine learning method can be used as the base-learner. In our experiments, we consider three widely used machine learning models including a two-layer \texttt{NN}, \texttt{SVR} and \texttt{LSTM}, for a proof of concept purpose.
            \item hyperparameters associated with the base-learner: the number of hidden neurons of \texttt{NN} (from $128$ to $1,024$), kernels used in \texttt{SVR} (including linear, poly, rbf, sigmoid, precomputed), and the number of units in \texttt{LSTM} (from $128$ to $1,024$).
            \item Three different learning rates $\alpha\in[0.0001,0.5]$, $\beta\in[0.0001,0.5]$, and $\gamma\in[0.0001,0.5]$.
            \item Optimization methods for the loss function: SGD~\cite{Amari93}, Adam~\cite{KingmaB17}, RMSprop~\cite{SaadnaBA21}, Adadelta~\cite{Zeiler12} and Adagrad~\cite{TraoreP21}.
        \end{itemize}
    
    \item\underline{\textit{Performance metric}}: Here we use the widely used mean squared error (MSE) as the metric to evaluate the predictive accuracy of a forecasting model.
        \begin{equation}
            \mathrm{MSE}=\frac{1}{N}\sum_{i=1}^{N}(y_i-\hat{y_i})^2,
            \label{eq:mse}
        \end{equation}
    where $N$ is number of instances in the testing set, $y_i$ and $\hat{y_i}$ are the ground truth and predicted value, respectively.

    \item\underline{\textit{Statistical tests}}: For statistical interpretation of the significance of the comparison results, we apply the following two statistical methods in our empirical study.
        \begin{itemize}
	        \item\underline{Wilcoxon signed-rank test}~\cite{Wilcoxon1992}: It is a non-parametric statistical test that makes no assumption about the underlying distribution of the data. In particular, the significance level is set to $p=0.05$ in our experiments.
            \item\underline{$A_{12}$ effect size}~\cite{LiC21}: To ensure the resulted differences are not generated from a trivial effect, we apply $A_{12}$ as the effect size measure to evaluate the probability that one algorithm is better than another. Specifically, given a pair of peer algorithms, $A_{12}=0.5$ means they are \textit{equivalent}. $A_{12}>0.5$ denotes that one is better for more than 50\% of the times. $0.56\leq A_{12}<0.64$ indicates a \textit{small} effect size while $0.64 \leq A_{12} < 0.71$ and $A_{12} \geq 0.71$ mean a \textit{medium} and a \textit{large} effect size, respectively. 
    \end{itemize}
\end{itemize}

%!Tex root=main.tex

\section{Results and Discussions}
\label{sec:results}

Our empirical study is driven by addressing the following three research questions (RQs).
\begin{itemize}
    \item\underline{\textbf{RQ1}}: Does \our\ framework work for different types of energy sources?
    \item\underline{\textbf{RQ2}}: Does the meta-learning alone can help a base-learner handle small data challenge?
    \item\underline{\textbf{RQ3}}: What is the added value of involving different types of configuration options in the hyperparameter optimization of a few-shot learning pipeline?
    \item\underline{\textbf{RQ4}}: What is the benefits of the bi-level programming formulation in \our?
    \item\underline{\textbf{RQ5}}: What is the impact of the computational budget allocated to the upper-level optimization?
\end{itemize}

\subsection{Performance Evaluation of the Effectiveness of the Proposed \our\ Framework}
\label{sec:rq1}

\subsubsection{Methods}
\label{sec:methods_rq1}

As introduced in~\pref{sec:setup}, by using \texttt{NN}, \texttt{SVR}, and \texttt{LSTM} as the base-learners, we come up with three instances of our \our\ framework, denoted as \texttt{BiLO-Auto-TSF/ML-NN}, \texttt{BiLO-Auto-TSF/ML-SVR} and \texttt{BiLO-Auto-TSF/ML-LSTM} respectively. As introduced in~\pref{sec:setup}, the historical data of wind generation, PV generation, and load demand for a period of a week are used to constitute the training dataset while the performance of different forecasting models is validated on the island Lundy for a period of $24$ hours.

\subsubsection{Results}
\label{sec:results_rq1}

\begin{table}[t!]
    \scriptsize
    \centering
    \caption{Comparison results of the MSE values obtained by different models for forecasting tasks of wind generation, PV generation and load demand.}
    \begin{tabular}{|c|c|c|c|}
        \cline{2-4}
        \multicolumn{1}{c|}{} & \multicolumn{3}{c|}{Wind generation} \\\cline{2-4}
        \multicolumn{1}{c|}{} & $n^{\mathrm{g}}=1$     & $n^{\mathrm{g}}=2$     & $n^{\mathrm{g}}=10$ \\ \hline
        \texttt{NN}    & 1.041E-1(7.01E-3)$^{\dagger}$ & 9.453E-2(5.82E-4)$^{\dagger}$ & 7.845E-2(4.42E-4)$^{\dagger}$ \\ \hline
        \texttt{ML-NN} & 5.922E-2(3.56E-3)$^{\dagger}$ & 5.33E-2(5.71W-4)$^{\dagger}$ & 3.925E-2(8.38E-5)$^{\dagger}$ \\ \hline
        \texttt{BiLO$^\ast$-NN} & 5.248E-2(7.32E-5)$^{\dagger}$ & 4.736E-2(2.52E-6)$^{\dagger}$ & 3.624E-2(4.82E-6)$^{\dagger}$ \\ \hline\hline
        \texttt{SVR}   & 1.082E-1(1.84E-3)$^{\dagger}$ & 9.845E-2(8.12E-4)$^{\dagger}$ & 7.919E-2(1.12E-4)$^{\dagger}$ \\ \hline
        \texttt{ML-SVR} & 6.017E-2(1.32E-4)$^{\dagger}$ & 5.761E-2(4.74E-4)$^{\dagger}$ & 4.233E-2(5.71E-5)$^{\dagger}$ \\ \hline
        \texttt{BiLO$^\ast$-SVR} & 5.217E-2(6.69E-5)$^{\dagger}$ & 4.754E-2(2.62E-6)$^{\dagger}$ & 3.546E-2(3.59E-6)$^{\dagger}$ \\ \hline\hline
        \texttt{LSTM}  & 7.684E-2(6.82E-3)$^{\dagger}$ & 6.892E-2(9.72E-4)$^{\dagger}$ & 5.622E-2(3.73E-4)$^{\dagger}$ \\
        \hline
        \texttt{ML-LSTM} & 5.247E-2(8.25E-4)$^{\dagger}$ & 4.818E-2(7.78E-4)$^{\dagger}$ & 3.232E-2(3.58E-5)$^{\dagger}$ \\
        \hline
        \texttt{BiLO$^\ast$-LSTM} & \multicolumn{1}{>{\columncolor{mycyan}}c|}{\textbf{4.125E-2(9.55E-5)}} & \multicolumn{1}{>{\columncolor{mycyan}}c|}{\textbf{3.844E-2(6.87E-6)}} & \multicolumn{1}{>{\columncolor{mycyan}}c|}{\textbf{2.476E-2(8.55E-6)}} \\ \hline
        
        \multicolumn{1}{c|}{} & \multicolumn{3}{c|}{PV generation} \\\cline{2-4}
        \multicolumn{1}{c|}{} & $n^{\mathrm{g}}=1$     & $n^{\mathrm{g}}=2$     & $n^{\mathrm{g}}=10$ \\ \hline
        \texttt{NN}    & 1.313E-1(2.42E-2)$^{\dagger}$ & 1.157E-1(1.92E-2)$^{\dagger}$ & 1.049E-1(1.15E-2)$^{\dagger}$ \\ \hline
        \texttt{ML-NN} & 3.621E-2(1.25E-3)$^{\dagger}$ & 2.972E-2(4.82E-4)$^{\dagger}$ & 2.238E-2(7.48E-5)$^{\dagger}$ \\ \hline
        \texttt{BiLO$^\ast$-NN} & 2.592E-2(3.72E-5)$^{\dagger}$ & 2.276E-2(1.26E-6)$^{\dagger}$ & 1.785E-2(5.34E-6)$^{\dagger}$ \\ \hline\hline
        \texttt{SVR}   & 1.261E-1(3.42E-2)$^{\dagger}$ & 1.077E-1(5.67E-2)$^{\dagger}$ & 9.854E-2(4.39E-2)$^{\dagger}$ \\ \hline
        \texttt{ML-SVR} & 3.443E-2(3.11E-3)$^{\dagger}$ & 2.823E-2(7.25E-4)$^{\dagger}$ & 2.107E-2(4.65E-5)$^{\dagger}$ \\ \hline
        \texttt{BiLO$^\ast$-SVR} & 2.318E-2(1.72E-5)$^{\dagger}$ & 2.179E-2(3.85E-5)$^{\dagger}$ & 1.762E-2(5.36E-6)$^{\dagger}$ \\ \hline\hline
        \texttt{LSTM}  & 5.482E-2(1.01E-3)$^{\dagger}$ & 4.743E-2(7.72E-3)$^{\dagger}$ & 3.934E-2(8.25E-3)$^{\dagger}$ \\ \hline
        \texttt{ML-LSTM} & 3.122E-2(6.23E-3)$^{\dagger}$ & 2.173E-2(8.45E-4)$^{\dagger}$ & 1.775E-2(5.58E-5)$^{\dagger}$ \\ \hline
        \texttt{BiLO$^\ast$-LSTM} & \multicolumn{1}{>{\columncolor{mycyan}}c|}{\textbf{2.215E-2(6.54E-6)}} & \multicolumn{1}{>{\columncolor{mycyan}}c|}{\textbf{1.943E-2(7.25E-6)}}  & \multicolumn{1}{>{\columncolor{mycyan}}c|}{\textbf{1.372E-2(8.66E-6)}}  \\ \hline
        
        \multicolumn{1}{c|}{} & \multicolumn{3}{c|}{Load demand} \\ \cline{2-4}
        \multicolumn{1}{c|}{} & $n^{\mathrm{g}}=1$     & $n^{\mathrm{g}}=2$     & $n^{\mathrm{g}}=10$ \\ \hline
        \texttt{NN}    & 7.518E-2(3.52E-3)$^{\dagger}$ & 7.214E-2(2.67E-3)$^{\dagger}$ & 6.278E-2(1.78E-4)$^{\dagger}$ \\ \hline
        \texttt{ML-NN} & 6.338E-2(3.12E-3)$^{\dagger}$ & 5.682E-2(6.45E-4)$^{\dagger}$ & 5.017E-2(8.62E-5)$^{\dagger}$ \\ \hline
        \texttt{BiLO$^\ast$-NN} & 5.231E-2(6.54E-5)$^{\dagger}$ & 4.776E-2(9.53E-5)$^{\dagger}$ & 3.890E-2(3.42E-6)$^{\dagger}$ \\ \hline\hline
        \texttt{SVR}   & 6.945E-2(4.01E-3)$^{\dagger}$ & 6.522E-2(6.84e-4)$^{\dagger}$ & 6.116E-2(4.12E-4)$^{\dagger}$ \\ \hline
        \texttt{ML-SVR} & 6.229E-2(2.23E-3)$^{\dagger}$ & 5.943E-2(3.67E-4)$^{\dagger}$ & 5.334E-2(5.66E-5)$^{\dagger}$ \\ \hline
        \texttt{BiLO$^\ast$-SVR} & 5.317E-2(2.42E-5)$^{\dagger}$ & 4.882E-2(6.76E-6)$^{\dagger}$ & 3.631E-2(7.66E-6)$^{\dagger}$ \\ \hline\hline
        \texttt{LSTM}  & 6.853E-2(1.71E-3)$^{\dagger}$ & 6.659E-2(4.58E-4)$^{\dagger}$ & 4.547E-2(5.87E-4)$^{\dagger}$ \\ \hline
        \texttt{ML-LSTM} & 4.828E-2(1.04E-3)$^{\dagger}$ & 4.113E-2(5.75E-4)$^{\dagger}$ & 3.282E-2(3.42E-6)$^{\dagger}$ \\ \hline
        \texttt{BiLO$^\ast$-LSTM} & \multicolumn{1}{>{\columncolor{mycyan}}c|}{\textbf{4.254E-2(3.65E-6)}} & \multicolumn{1}{>{\columncolor{mycyan}}c|}{\textbf{3.123E-2(5.72E-6)}}  & \multicolumn{1}{>{\columncolor{mycyan}}c|}{\textbf{2.524E-2(6.88E-6)}} \\ \hline
    \end{tabular}
    \begin{tablenotes}
        \footnotesize
            \item[1] $^{\dagger}$ indicates the better result (highlighted in bold face with a gray background) is of statistical significance according to the Wilcoxon's rank-sum test at the $5\%$ significance level.
    \end{tablenotes}
    \label{tab:overall_performance}%
\end{table}%

\begin{figure*}[t!]
\centering
\includegraphics [width=\linewidth]{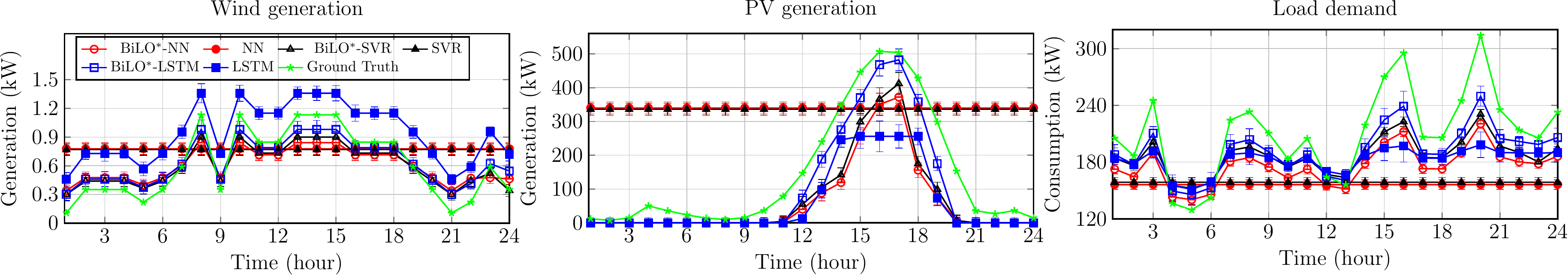}
\caption{Comparison of the forecasting performance of different models for wind generation, PV generation and load demand, respectively. Note that BiLO$^\ast$ is short for \our.}
\label{fig:comparison_three_models}
\end{figure*}

From the comparison results shown in~\pref{tab:overall_performance}, we can see that the performance of all three base-learners have been significantly improved by being embedded into our proposed \our\ framework. To have a better visual interpretation of the performance comparison, we plot the forecasting results of $24$ hours for wind generation, PV generation and load demand in~\pref{fig:comparison_three_models}. From these trajectories, it is clear to see that the vanilla \texttt{NN} and \texttt{SVR} cannot make any meaningful forecasting at all in all cases. This is anticipated as the training data at hand is extremely scarce so that neither \texttt{NN} nor \texttt{SVR} can be appropriately trained. Although the vanilla \texttt{LSTM} can capture the variation of the time series data as shown in~\pref{fig:comparison_three_models}, its prediction is largely offset with regard to the ground truth. In contrast, by being equipped with our proposed \our, the performance of all base-learners have been significantly improved. Especially for \texttt{NN} and \texttt{SVR}, after using few-shot learning and hyperparameter optimization, the performance of \texttt{BiLO-Auto-TSF/ML-NN} and \texttt{BiLO-Auto-TSF/ML-SVR} have been leveled up to become comparable with \texttt{BiLO-Auto-TSF/ML-LSTM}.

\vspace{0.5em}
\noindent
\framebox{\parbox{\dimexpr\linewidth-2\fboxsep-2\fboxrule}{
        \textbf{\underline{Response to RQ1:}} \textit{From the observations in this experiment, it is clear to see that our proposed \our\ framework is able to significantly improve the forecasting performance of a base-learner for different types of energy source. In particular, the adaptation to extremely scarce historical data can be attributed to the few-shot learning in \our.}
    }}

% Table I lists the fine-tuning test results of three machine learning models trained by standard training strategy, meta-learning and \our, respectively. The best result of each type of data is marked in bold font. Compared to the results of the NN model, SVR model and LSTM model with standard training strategy, these machine learning models with \our\ can achieve better performance on all types of data.

\subsection{Investigation of the Effectiveness of Meta-Learning}
\label{sec:rq2}

\subsubsection{Methods}
\label{sec:methods_rq2}

The results in~\pref{sec:rq1} have shown the overall superiority of our proposed \our\ framework for improving a base-learner for time series forecasting with extremely limited historical data. To address \textbf{RQ2}, we directly apply the meta-learning approach introduced in~\pref{sec:lower} to each of \texttt{NN}, \texttt{SVR} and \texttt{LSTM} models. The resulted forecasting models are denoted as \texttt{ML-NN}, \texttt{ML-SVR} and \texttt{ML-LSTM}, respectively. Note that the hyperparameters associated with the base- and meta-learners are fixed a priori (e.g., we set the number of neurons in the \texttt{NN} to be $512$, the kernel is set to be rbf, and the number of cells in the \texttt{LSTM} to $640$, three different learning rates set as $0.01$, $0.001$, and $0.05$, respectively, and SGD is selected as the optimizer). In addition, we plan to investigate the influence of the number of gradient descent steps (denoted as $n^{\mathrm{g}}$) for fine-tuning the updated parameters $\hat{\boldsymbol{\theta}}$ during the meta training phase. Specifically, $n^{\mathrm{g}}$ is set as $1$, $2$, and $10$ in this experiment.

\subsubsection{Results}
\label{sec:results_rq2}

\begin{figure}[t!]
    \centering
    \includegraphics [width=.5\linewidth]{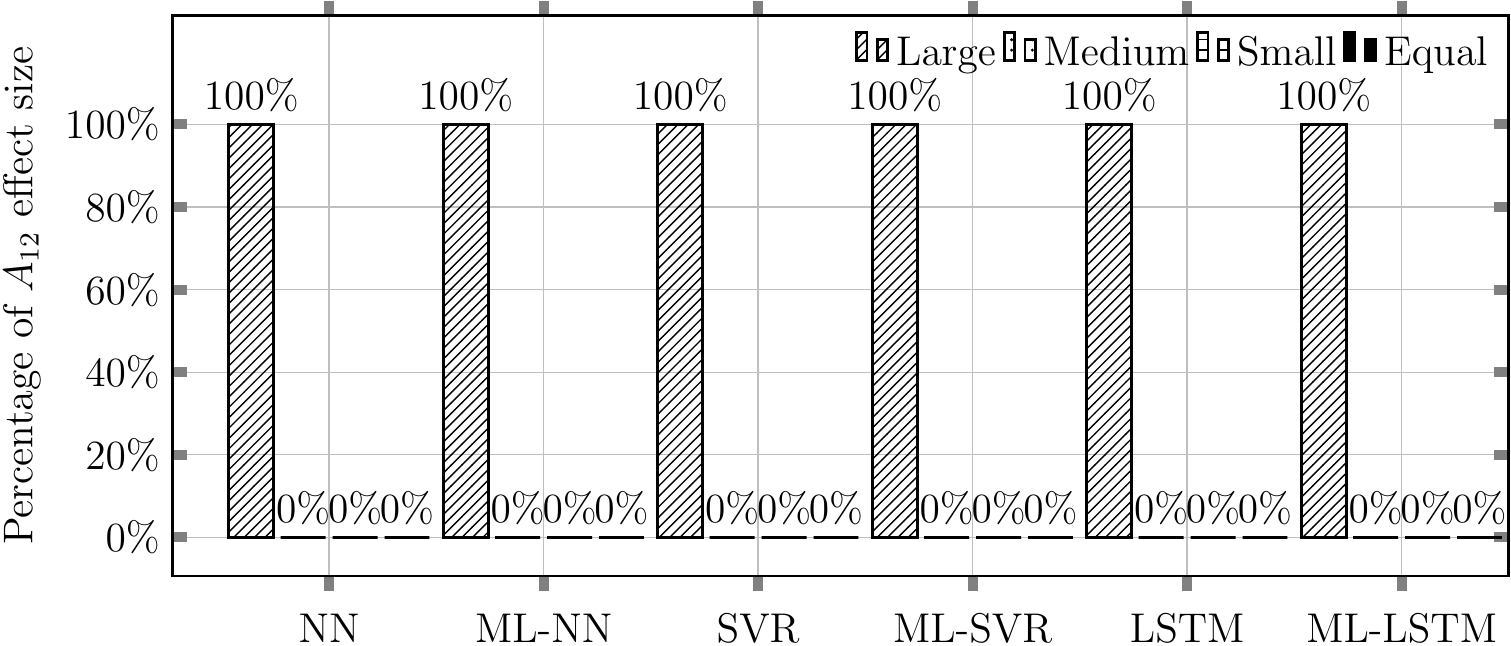}
    \caption{Percentage of the large, medium, small, and equal $A_{12}$ effect size, respectively, when comparing our proposed \texttt{BiLO-Auto-TSF/ML} framework based models against their base-learners and the one only using meta-learning.}
    \label{fig:a12_comp_ours}
\end{figure}

From the comparison results shown in~\pref{tab:overall_performance}, we find that the performance of the vanilla \texttt{NN}, \texttt{SVR}, and \texttt{LSTM} can be improved by directly using the meta-learning. In particular, it is worth noting that the corresponding hyperparameters of \texttt{ML-NN}, \texttt{ML-SVR}, and \texttt{ML-LSTM} are fixed a priori whereas our proposed \our\ framework is able to automatically search for an optimal few-shot learning pipeline for the underlying base-learner. As the comparison results shown in~\pref{tab:overall_performance}, we find that the performance of \texttt{ML-NN}, \texttt{ML-SVR}, and \texttt{ML-LSTM} can be further improved under our proposed \our\ framework. To have a better investigation for the performance difference achieved by using \our\ framework against its vanilla version and the conventional meta-learning, we show the statistical results of $A_{12}$ in~\pref{fig:a12_comp_ours}. From this result, it is clear to see that the better performance achieved by using \our\ framework are always classified to have a large effect size. This observation supports the importance of choosing appropriate hyperparameters in time series forecasting for a given dataset.

\begin{figure*}[t!]
    \centering
    \includegraphics [width=\linewidth]{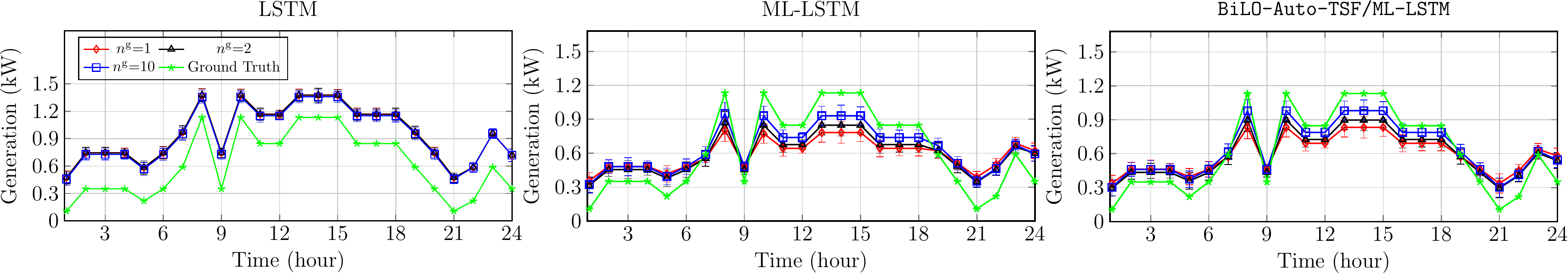}
    \caption{Comparison of the forecasting performance of \texttt{LSTM}, \texttt{ML-LSTM} and \texttt{BiLO-Auto-TSF/ML-LSTM} with different number of gradient descent steps for the wind generation task.}
    \label{fig:lstm_wind}
\end{figure*}

\begin{figure*}[t!]
    \centering
    \includegraphics[width=\linewidth]{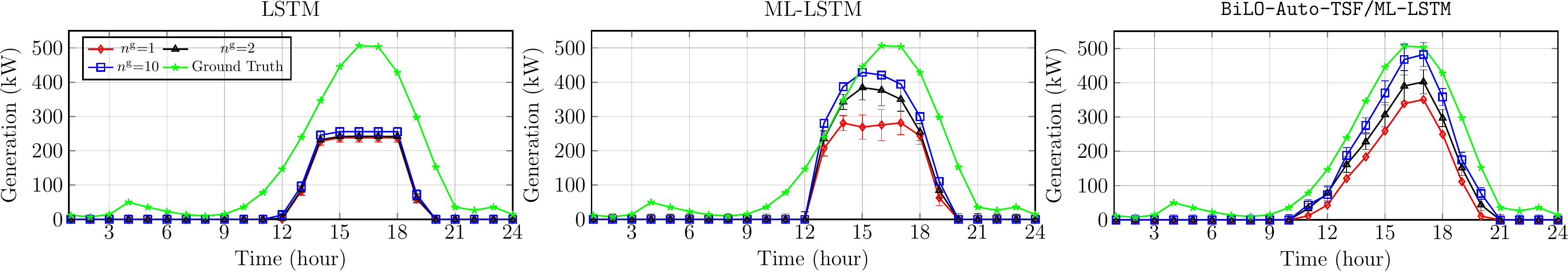}
    \caption{Comparison of the forecasting performance of \texttt{LSTM}, \texttt{ML-LSTM} and \texttt{BiLO-Auto-TSF/ML-LSTM} with different number of gradient descent steps for the PV generation task.}
    \label{fig:lstm_pv}
\end{figure*}

\begin{figure*}[t!]
    \centering
    \includegraphics [width=\linewidth]{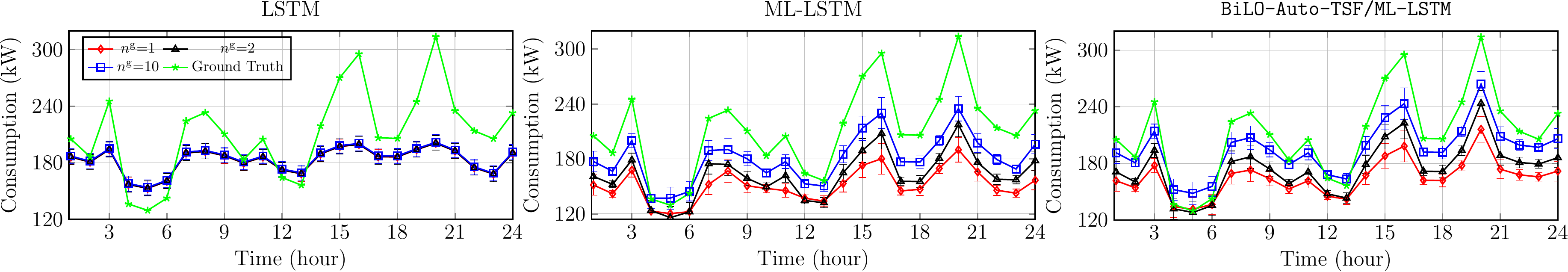}
    \caption{Comparison of the forecasting performance of \texttt{LSTM}, \texttt{ML-LSTM} and \texttt{BiLO-Auto-TSF/ML-LSTM} with different number of gradient descent steps for the load demand task.}
    \label{fig:lstm_load}
\end{figure*}

Furthermore, let us look into the influence of the number of gradient descent steps. As the results shown in~\pref{tab:overall_performance}, it is appreciated that the forecasting performance can be further improved by increasing the number of gradient descent steps during the meta training phase. This is not difficult to understand as more gradient descent steps lead to a better fine-tuned result. To have a visual illustration of the impact of $n^{\mathrm{g}}$, we plot the forecasting results of \texttt{LSTM}, \texttt{ML-LSTM} and \texttt{BiLO-Auto-TSF/ML-LSTM} for different energy sources in Figs.~\ref{fig:lstm_wind} to~\ref{fig:lstm_load}. From these trajectories, it is clear to see that the number of gradient descent steps does not have any visible impact on the performance of the vanilla \texttt{LSTM}; whereas it is able to fine tune the performance of meta-learning.

\begin{figure}[t!]
    \centering
    \includegraphics [width=.7\linewidth]{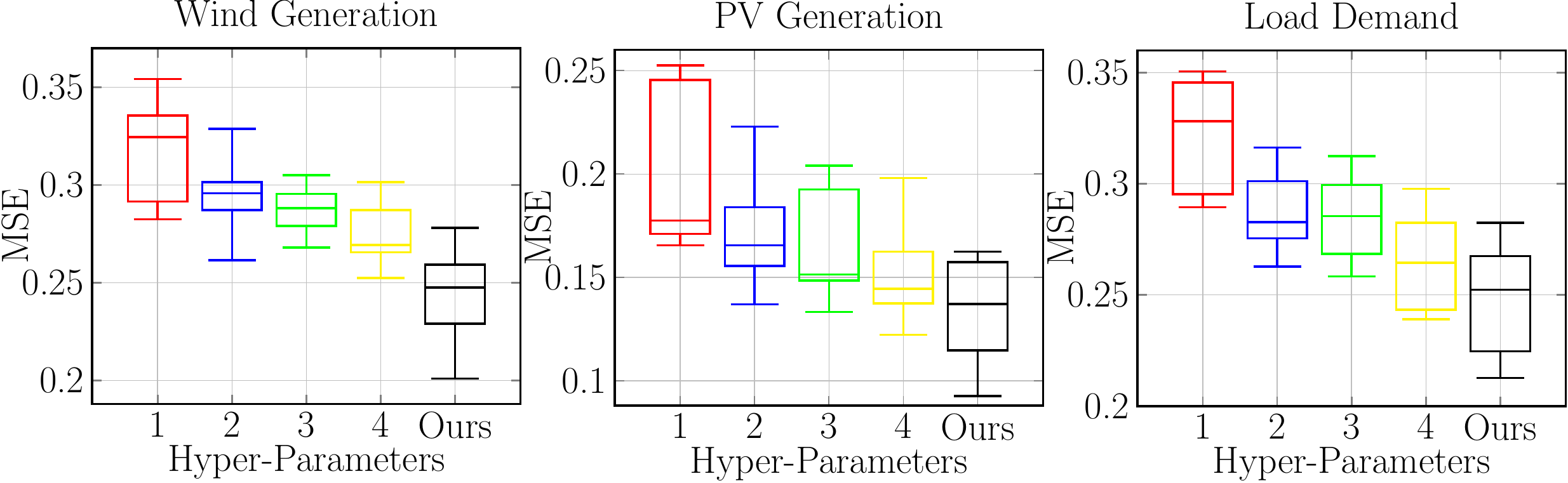}
    \caption{Comparison of the forecasting performance of \texttt{BiLO-Auto-TSF/ML-LSTM} against the other the variants \our\ that involve different number of hyperparameters in optimization, represented as the index $1\leq i<5$ in the $x$-coordinate.}
    \label{fig:boxplot_diff_num}
\end{figure}

% \vspace{0.5em}
\noindent
\framebox{\parbox{\dimexpr\linewidth-2\fboxsep-2\fboxrule}{
    \textbf{\underline{Response to RQ2:}} \textit{There are three takeaways from this experiment. First, meta-learning can enable a vanilla machine learning model to be capable of carrying out time series forecasting with extremely limited historical data. Second, by using our proposed \our\ framework, the performance of the corresponding forecasting model can be further improved. This can be attributed to the hyperparameter optimization that helps identifies the most competitive few-shot learning pipeline. Last but not the least, using more gradient descent steps can be beneficial for fine tuning during the meta-training phase.}
}}

\begin{figure}[htbp]
    \centering
    \includegraphics [width=.3\linewidth]{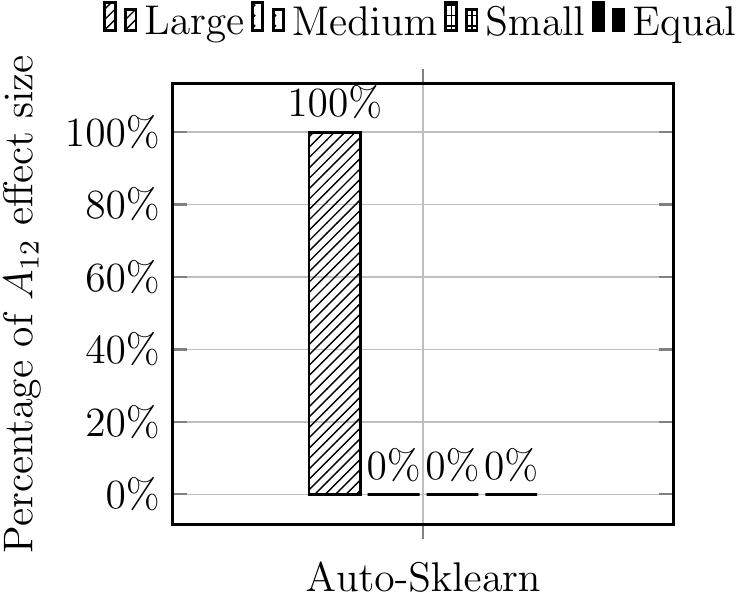}
    \caption{Percentage of the large, medium, small, and equal $A_{12}$ effect size, respectively, when comparing our proposed \texttt{BiLO-Auto-TSF/ML} framework based models against Auto-Sklearn framework.}
    \label{fig:a12_comp_sk}
\end{figure}

\subsection{Investigation of the Hyperparameters Optimization for Meta-Learning}
\label{sec:rq3}

\subsubsection{Methods}
\label{sec:methods_rq3}

As introduced in~\pref{sec:setup}, there are more than one type of hyperparameters considered in our few-shot learning pipeline. Another question is whether we need to optimize all those hyperparameters or can we achieve comparable performance by only optimizing some of them? To address \textbf{RQ3}, we come up with $\sum_{i=1}^4 {{5}\choose{i}}$ different variants by considering $1\leq i<5$ hyperparameters in \our. Given the outstanding performance observed in~\pref{sec:results_rq1}, here we only consider \texttt{LSTM} as the base-learner in this experiment without loss of generality. To address \textbf{RQ4}, we compare the performance of \our\ with \texttt{Auto-Sklearn}~\cite{FeurerEFLH20}, one of the most popular tools in the automated machine learning literature. Here we still only consider \texttt{LSTM} as the base-learner as before without loss of generality. Note that \texttt{Auto-Sklearn} does not apply a bilevel programming paradigm for hyperparameter optimization.

\subsubsection{Results}
\label{sec:results_rq3}

\begin{table}[t!]
  \centering
  \caption{Comparison results of MSE values obtained by \texttt{Auto-Sklearn} and \texttt{BiLO-Auto-TSF/ML-LSTM} }
    \begin{tabular}{ccc}
\cline{2-3}    \multicolumn{1}{c|}{} & \multicolumn{2}{c|}{MSE} \\
    \hline
    \multicolumn{1}{|c|}{\texttt{Data}} & \multicolumn{1}{c|}{\texttt{Auto-Sklearn}} & \multicolumn{1}{c|}{\texttt{BiLO$\ast$-LSTM}} \\
    \hline
    \multicolumn{1}{|c|}{Wind generation} & \multicolumn{1}{c|}{5.143E-2(4.76E-6)} & \multicolumn{1}{>{\columncolor{mycyan}}c|}{\textbf{2.476E-2(8.55E-6)}} \\
    \hline
    \multicolumn{1}{|c|}{PV generation} & \multicolumn{1}{c|}{1.745E-2(6.42E-6)} & \multicolumn{1}{>{\columncolor{mycyan}}c|}{\textbf{1.372E-2(8.66E-6)}} \\
    \hline
    \multicolumn{1}{|c|}{Load demand} & \multicolumn{1}{c|}{3.628E-2(2.93E-6)} & \multicolumn{1}{>{\columncolor{mycyan}}c|}{\textbf{2.524E-2(6.88E-6)}} \\ \hline
    \end{tabular}%
  \label{tab:comp_sk}%
\end{table}%

From the box-plots of the MSE values obtained by different variants of \our\ considering $1\leq i<5$ hyperparameters shown in~\pref{fig:boxplot_diff_num}, it is clear to see that our proposed \our\ constantly outperforms the other variants. It is interesting to note that the predictive accuracy can be improved by involving more types of hyperparameters in the optimization. In other words, we can envisage a further improvement of \our\ by involving more configuration options in the few-shot learning pipeline.

According to the comparison results shown in~\pref{tab:comp_sk}, we can see that the performance of \our\ is constantly better than that of \texttt{Auto-Sklearn} in all three types of forecasting tasks. Furthermore, as the $A_{12}$ results shown in~\pref{fig:a12_comp_sk}, it is clear to see that the better results achieved by our proposed \our\ is always classified to have a large effect size. It is worth noting that one of the key differences between \our\ and \texttt{Auto-Sklearn} lies in the bi-level programming perspective for coordinating the meta-learning and hyperparameter optimization in an intertwined manner. 

\noindent
\framebox{\parbox{\dimexpr\linewidth-2\fboxsep-2\fboxrule}{
        \textbf{\underline{Response to RQ3:}} \textit{From the observations in this experiment, we can see that the performance of a few-shot learning pipeline can be improved by involving more configuration options in the hyperparameter optimization.
        }
}}

\vspace{0.2em}
\noindent
\framebox{\parbox{\dimexpr\linewidth-2\fboxsep-2\fboxrule}{
       \textbf{\underline{Response to RQ4:}} \textit{From the comparison results w.r.t. \texttt{Auto-Sklearn}, we confirm the effectiveness of bi-level programming paradigm for handling meta-learning and hyperparameter optimization in a concurrent manner.
    }
}}

% \vspace{2cm}
% \noindent
% \framebox{\parbox{\dimexpr\linewidth-2\fboxsep-2\fboxrule}{
%         \textit{ically, this improvement can be attributed to the optimized hyper-parameters combination by using our proposed \our\ framework.}
% }}

\subsection{Impact of Computational Budget at the Upper Level}
\label{sec:rq4}

\subsubsection{Methods}
\label{sec:methods_rq4}

In practice, it is not uncommon that the computational resource for time series forecasting, in particular the time budget, is limited. Given the iterative nature of MCTS, searching for the optimal few-shot learning pipeline at the upper-level optimization by using a MCTS can be time consuming. In this subsection, we are interested to investigate how is the number of iterations in MCTS (i.e., the computational budget allocated to the upper-level optimization) related to the performance. To this end, for each of the three instances of our proposed \our\ framework, we keep a record of the variation of the MSE across $720$ iterations in MCTS for the forecasting tasks of wind generation, PV generation and load demand, respectively.
% The results in~\pref{sec:rq3} have shown the performance of hyperparameters optimization by being embedded into the bi-level programming formulation. To identify the benefits of bi-level programming formulation, we investigate the convergence of MCTS method for wind generation, PV generation, and load demand. Specifically, the maximum iterations is set as 720 in this experiment.

\subsubsection{Results}
\label{sec:results_rq4}

\begin{figure*}[t!]
    \centering
    \includegraphics [width=\linewidth]{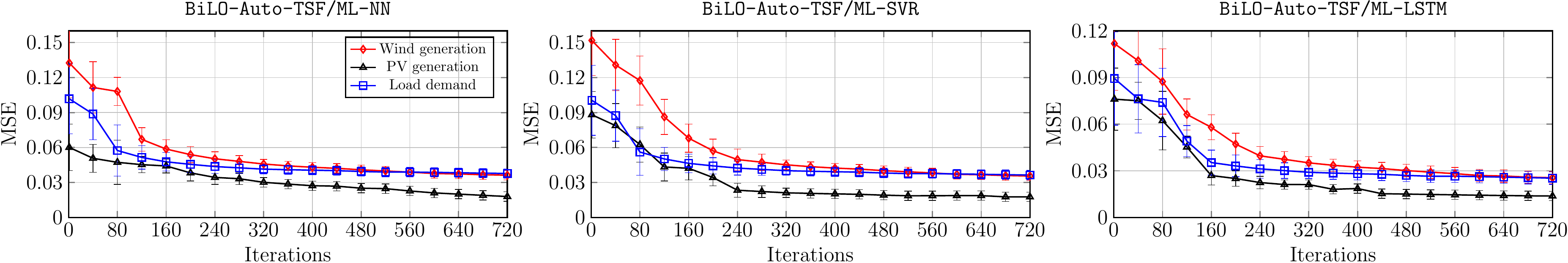}
    \caption{The trajectories of the MSE across $720$ iterations in the MCTS of \our\texttt{-NN}, \our\texttt{-SVR} and \our\texttt{-LSTM}.}
    \label{fig:mcts_traj}
\end{figure*}

\begin{figure*}[t!]
    \centering
    \includegraphics [width=\linewidth]{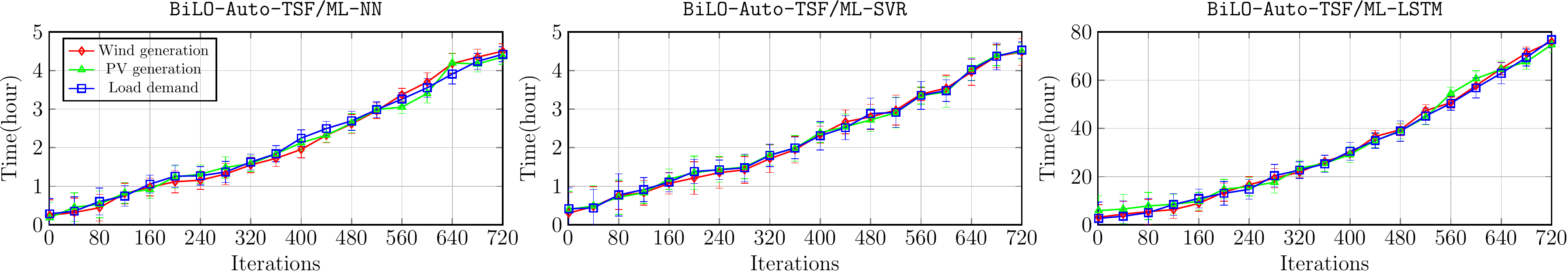}
    \caption{The trajectories of the CPU wall clock time across $720$ iterations in the MCTS of \our\texttt{-NN}, \our\texttt{-SVR} and \our\texttt{-LSTM}.}
    \label{fig:mcts_time}
\end{figure*}

From the trajectories shown in~\pref{fig:mcts_traj}, we can see that the overall MSE keeps on reducing with the increase of the iterations in MCTS. However, for all three \our\ instances, the MSE trajectories does not show to be significantly changed after around the $240$-th iteration. This suggests that we can reduce the number of iterations in MCTS without significantly deteriorating the performance of the identified few-shot learning pipeline. In addition, as the trajectories of CPU wall clock time shown in~\pref{fig:mcts_time}, we can see that the computational time can be significantly reduced when early terminating the upper-level optimization routine.

\vspace{0.5em}
\noindent
\framebox{\parbox{\dimexpr\linewidth-2\fboxsep-2\fboxrule}{
        \textbf{\underline{Response to RQ5:}} \textit{From the experiment in this subsection, we find that the computational budget allocated to the upper-level hyperparameter optimization of the few-shot learning pipeline can be narrowed down without significantly compromising the forecasting performance.}
    }}

%!TeX root=main.tex

\section{Conclusion}
\label{sec:conclusion}

Short- and/or long-term time series forecasting of both energy generation and load demand have been one of the key tools to guide the optimal decision-making for planning and operation of utility companies without over/underestimating the capabilities of renewable energy infrastructures. Different from the traditional big data scenarios, one of the most challenging issues in time series renewable energy forecasting in the real world is the short of historical data. This can render most prevalent machine learning models ineffective. In addition, the performance of machine learning models are sensitive to the choice of their corresponding hyperparameters w.r.t. the characteristics of the underlying forecasting tasks. Bearing these considerations in mind, this paper developed a \our\ framework that automatically searches for a high-performance few-shot learning pipeline from a bi-level programming perspective. More specifically, the meta-learning routine at the lower level helps mitigate the small data challenges while the hyperparameter optimization at the upper level helps search for the optimal configuration options to achieve the peak few-shot learning performance. Extensive experiments fully demonstrate the effectiveness of our proposed \our\ framework to significantly boost the performance of three prevalent machine learning models for time series renewable energy forecasting with extremely limited historical data.

\section*{Acknowledgment}
K. Li was supported by UKRI Future Leaders Fellowship (MR/S017062/1), Amazon Research Awards, Royal Society International Exchange Scheme (IES/R2/212077), Alan Turing Fellowship, and EPSRC (2404317).

\bibliographystyle{IEEEtran}
\bibliography{IEEEabrv, reference}
% Please fill your bibtex file name in 'xxx'.

\end{document}